\documentclass[iicol,sn-basic,Numbered]{sn-jnl}

\usepackage{graphicx}%
\usepackage{multirow}%
\usepackage{amsmath,amssymb,amsfonts}%
\usepackage{amsthm}%
\usepackage{mathrsfs}%
\usepackage[title]{appendix}%
\usepackage{xcolor}%
\usepackage{textcomp}%
\usepackage{manyfoot}%
\usepackage{booktabs}%
\usepackage{algorithm}%
\usepackage{algorithmicx}%
\usepackage{algpseudocode}%
\usepackage{listings}%
\begin{document}

\title[Detecting and recognizing characters in Greek papyri with YOLOv8, DeiT and SimCLR]{Detecting and recognizing characters in Greek papyri with YOLOv8, DeiT and SimCLR}

%%=============================================================%%
%% GivenName	-> \fnm{Joergen W.}
%% Particle	-> \spfx{van der} -> surname prefix
%% FamilyName	-> \sur{Ploeg}
%% Suffix	-> \sfx{IV}
%% \author*[1,2]{\fnm{Joergen W.} \spfx{van der} \sur{Ploeg} 
%%  \sfx{IV}}\email{iauthor@gmail.com}

%%=============================================================%%

\author*[1]{\fnm{Robert} \sur{Turnbull} \orcid{0000-0003-1274-6750}}\email{robert.turnbull@unimelb.edu.au}

\author[2]{\fnm{Evelyn} \sur{Mannix} \orcid{0000-0003-1798-6470}}\email{evelyn.mannix@unimelb.edu.au}

\affil*[1]{\orgdiv{Melbourne Data Analytics Platform}, \orgname{The University of Melbourne}, \orgaddress{\street{Grattan St.}, \city{Parkville}, \postcode{3010}, \state{VIC}, \country{Australia}}}

\affil[2]{\orgdiv{Melbourne Centre for Data Science}, \orgname{The University of Melbourne}, \orgaddress{\street{Grattan St.}, \city{Parkville}, \postcode{3010}, \state{VIC}, \country{Australia}}}

\abstract{\textbf{Purpose:} The capacity to isolate and recognize individual characters from facsimile images of papyrus manuscripts yields rich opportunities for digital analysis. For this reason the `ICDAR 2023 Competition on Detection and Recognition of Greek Letters on Papyri' was held as part of the 17\textsuperscript{th} International Conference on Document Analysis and Recognition. This paper discusses our submission to the competition.

\textbf{Methods:}  We used an ensemble of YOLOv8 models to detect and classify individual characters and employed two different approaches for refining the character predictions, including a transformer based DeiT approach and a ResNet-50 model trained on a large corpus of unlabelled data using SimCLR, a self-supervised learning method.

\textbf{Results:}  Our submission won the recognition challenge with a mAP of 42.2\%, and was runner-up in the detection challenge with a mean average precision (mAP) of 51.4\%. At the more relaxed intersection over union threshold of 0.5, we achieved the highest mean average precision and mean average recall results for both detection and classification. 

\textbf{Conclusion:} The results demonstrate the potential for these techniques for automated character recognition on historical manuscripts. We ran the prediction pipeline on more than 4,500 images from the Oxyrhynchus Papyri to illustrate the utility of our approach, and we release the results publicly in multiple formats.}


%%\pacs[JEL Classification]{D8, H51}


\maketitle

%% main text
\section{Introduction}
\label{sec:intro}

The challenge of publishing editions of manuscripts from the ancient world is immense. There are countless documents in libraries, museums and monasteries that have yet to be edited or published in a digital form. For example, the Oxyrhynchus Papyri, discovered by Bernard P. Grenfell and Arthur S. Hunt, contains approximately half a million papyri and papyrus fragments \citep{parsons2007} only a small fraction of which have been published. 

Machine learning can assist in the tasks of editing and analyzing these collections. The ability to detect and recognize individual characters is especially useful and can be used for tasks such as assisting with approximate dating and identifying \textit{disiecta membra} written by the same scribe \citep{faigenbaum2022computational}. Automating the transcription of these documents also allows large corpora of scanned documents to be quickly searched for keywords or phrases of interest \citep{alabau2014iterative}.

To compare and evaluate methods for this task, Mathias Seuret and colleagues ran the `ICDAR 2023 Competition on Detection and Recognition of Greek Letters on Papyri' as part of the 17\textsuperscript{th} International Conference on Document Analysis and Recognition \citep{icdar}. The competition evaluated two tasks, one for detecting the bounding box for individual characters and the other for correctly recognizing the letter within each bounding box. This paper outlines our submission and the results of applying our method to published images of the Oxyrhynchus Papyri.

\section{Data}
\label{sec:data}

\subsection{Competition Dataset}

The training dataset provided for the competition consisted of 153 images from 108 Greek papyrus manuscripts which preserve text from the \textit{Iliad} of Homer. Bounding boxes were added and annotated with the Greek letter for each character present. Apostrophes and periods were also annotated but ignored in evaluating the results. We divided the training dataset into five cross-validation partitions. Multiple images present in the same source manuscript were included in the same partition. The test dataset included 34 images from 31 manuscripts. Each submission required a JSON file in the COCO format to be uploaded to the competition site on CodaLab \citep{codalab}.

\subsection{Supplemental Data}
To supplement the images in the training dataset, we included 4,533 images from volumes XV--LXXXII of the Oxyrhynchus Papyri (excluding volume XXVII). Images from these volumes included literary texts like in the competition dataset but it also included documentary texts. To help train models specifically for recognition, we used the second version of the AL-PUB dataset which contains 205,797 cropped images of individual characters taken from images in the Oxyrhynchus Papyri collection as part of the `Ancient Lives Project' \citep{swindall2021exploring}. These images were labeled with the Greek letter from each image. 

% We used the second version of this dataset which does not distinguish between the Sigma character ($\Sigma$) with the Lunate Sigma (C), which occurs much more commonly on papyrus.

\section{Method}
\label{sec:method}

We first trained models to perform both detection and recognition of characters and then models specifically for the recognition task.

\subsection{Detection and preliminary recognition}

For character detection and preliminary recognition, we used YOLO (You Only Look Once) \citep{YOLO}. The YOLO neural network model uses a series of convolutional layers to predict bounding boxes and probabilities for each categorical class simultaneously. Initial detections are made on coarse cells produced from the convolutions and, if the same object is predicted in multiple cells, they are corrected using non-maximal suppression. The release of YOLO9000 \citep{YOLO9000} and YOLOv3 \citep{YOLOv3} brought about multiple enhancements. In 2020, Glen Jocher released a PyTorch implementation of YOLOv3 which he named YOLOv5, published by Ultralytics \citep{YOLOv5}. Our models used Jocher's later version YOLOv8\citep{YOLOv8}. We used the `x' sized pretrained model which is reported to have achieved a mAP\textsuperscript{val} of 68.2 on the COCO val2017 dataset. We trained the model using the test dataset at three image resolutions: small (1280$\times$1280), medium (1600$\times$1600) and large (2048$\times$2048). Each YOLOv8 model was trained for 200 epochs and the weights from the epoch with the best result on the validation set were saved. The trained model was used to make predictions on the unannotated Oxyrhynchus Papyri images which were used as pseudo-labels  \citep{lee2013pseudo} and concatenated with the competition training set. The YOLOv8 models were then trained using the combined input from the competition dataset and the pseudo-labels.

\subsection{Recognition specific models}

To enhance the recognition performance of the YOLOv8 results, we trained additional specialized recognition models. This allowed us to incorporate additional labeled and unlabeled data, and to leverage self-supervised learning approaches.

In the first instance, we trained a self-supervised SimCLR model \citep{chen2020simple} on all of the available data containing cropped Greek characters. This included the competition dataset, the AL-PUB dataset, and the characters detected by the YOLOv8 models from the Oxyrhynchus Papyri. 

The SimCLR approach takes a mini-batch of images as usual, but augments each image to produce two views and employs a contrastive loss to teach the network to match each augmented view in the batch with its positive pair. On datasets such as ImageNet and CIFAR-10, augmentations including cropping, horizontal flipping, color-jitter and Gaussian blur are used to get best results \citep{chen2020simple}. 

Selecting appropriate augmentations for the problem is important for models trained using SimCLR to perform well because these influence the image features leveraged by the model to distinguish different images within the training process \citep{chen2021intriguing}. In the context of Greek character detection, augmentations such as cropping and horizontal flipping may change the character sufficiently to make it a different class. Accordingly we reduced the degree of cropping, removed the horizontal flips and did stronger color-jitter and blur.

After pretraining a ResNet-50 convolutional neural network on the data, we fine-tuned the model \citep{chen2020big} using the labels available in the competition and AL-PUB datasets using fivefold cross-validation.

In the second instance, we trained DeiT transformer models using transfer learning \citep{touvron2021training}. This was an entirely supervised process so we used the same labeled data as for the ResNet-50 SimCLR models with the same cross-validation splits. For this approach we also used less aggressive cropping during training, so that a significant portion of the character of interest would always be present during the training process.

\subsection{Ensembling}
\label{sec:ensembling}

For our final submission, we ensembled the results from our trained models. For character detection, we ensembled the results for the YOLOv8 trained models using Weighted Boxes Fusion \citep{solovyev2021weighted}. We then cropped the characters resulting from this procedure and used our recognition models to make predictions for each. We first ensembled the predictions for the ResNet-50 SimCLR and DeiT-small models separately, using a hard majority voting approach. We then ensembled the YOLOv8 classification with the ResNet-50 SimCLR and DeiT-small overall votes, by adjusting the bounding box confidence by the proportion of votes for each method for a particular character.

\begin{figure*}[ht!]
\includegraphics[width=\textwidth]{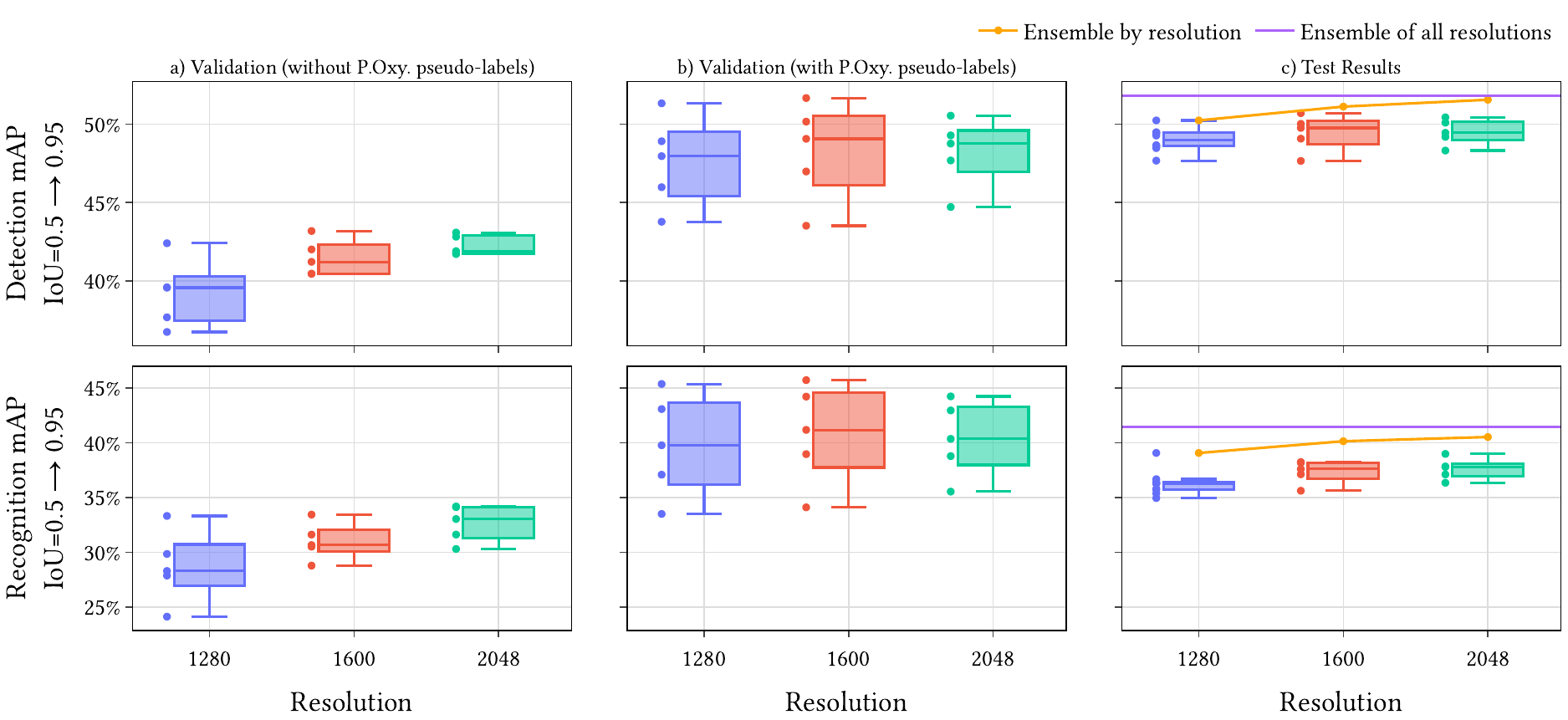}
\caption{YOLOv8 results. The boxplots show the five results trained with a different validation partition. The plots on the left show the results on the held-out validation set without using the pseudo-labels from the Oxyrhynchus Papyri. The plots in the center show the improved results on the respective validation sets with the pseudo-labels. The plots on the right show the results on the test dataset. The orange line represents an ensemble of the models at the three resolutions. The horizontal purple line shows the result of the ensemble of all trained models.}
\label{fig:yolov8-results}
\end{figure*}

\section{Evaluation}
\label{sec:evaluation}

The competition used the mean average precision (mAP) evaluation metric as used by the COCO dataset which averages results for ten intersection over union (IoU) values, with increments of 0.05 from 0.5 to 0.95. This was the primary metric which determined the winner for both the detection and recognition tasks. Results reported below use this metric unless stated otherwise. Also reported are the mean average precision and mean average recall (mAR) scores for an IoU value of 0.5, the traditional IoU value for evaluating object detection used in the \textsc{pascal} Visual Object Classes challenge \citep{everingham2010}, and an IoU of 0.75 which requires a stricter localization of the bounding box.

%All results below are with respect to test dataset and were computed on the CodaLab interface managed by the competition organizers.

\section{Results}
\label{sec:results}

\subsection{Experimental Results}
\label{sec:experimental-results}

Fig. \ref{fig:yolov8-results}a shows the mean average precision results of cross-validation on the training dataset. Increasing the resolution yielded a small improvement for both detection and recognition. But adding pseudo-labels (fig. \ref{fig:yolov8-results}b) from predictions of the model on the Oxyrhynchus Papyri images resulted in a 6--8\% increase for detection and a 8--11\% increase in recognition.

% \begin{figure}[ht!]
% \includegraphics[width=\columnwidth]{figs/YOLOv8-Resolutions.pdf}
% \caption{YOLOv8 test results. The boxplots show the results corresponding with the models trained with each time a different partition used for validation. The orange line represents an ensemble of the models at the three resolutions. The horizontal purple line shows the result of the ensemble of all trained models.}
% \label{fig:yolov8-results}
% \end{figure}

Fig. \ref{fig:yolov8-results}c shows results on the test set for YOLOv8 models trained on the five cross-validation partitions and at the three image resolutions. We achieved better results with higher resolutions but with diminishing returns. The average mAP score for detection across the five cross-validation partitions at the small resolution was 48.7\%. This increased to 49.5\% at the medium resolution but a further increase to the large resolution only raised the average mAP to 49.6\%. In terms of letter recognition, the lower resolution gave a mAP of 35.8\% while the medium and large resolutions gave 37.3\% and 37.6\% respectively. Ensembling the five models resulted in a distinct improvement for both detection and recognition. The final ensemble combining the YOLOv8 models at all resolutions produced the best results with a mAP score of 51.8\% for detection and 41.4\% for recognition.

\begin{figure*}[ht!]
\includegraphics[width=\textwidth]{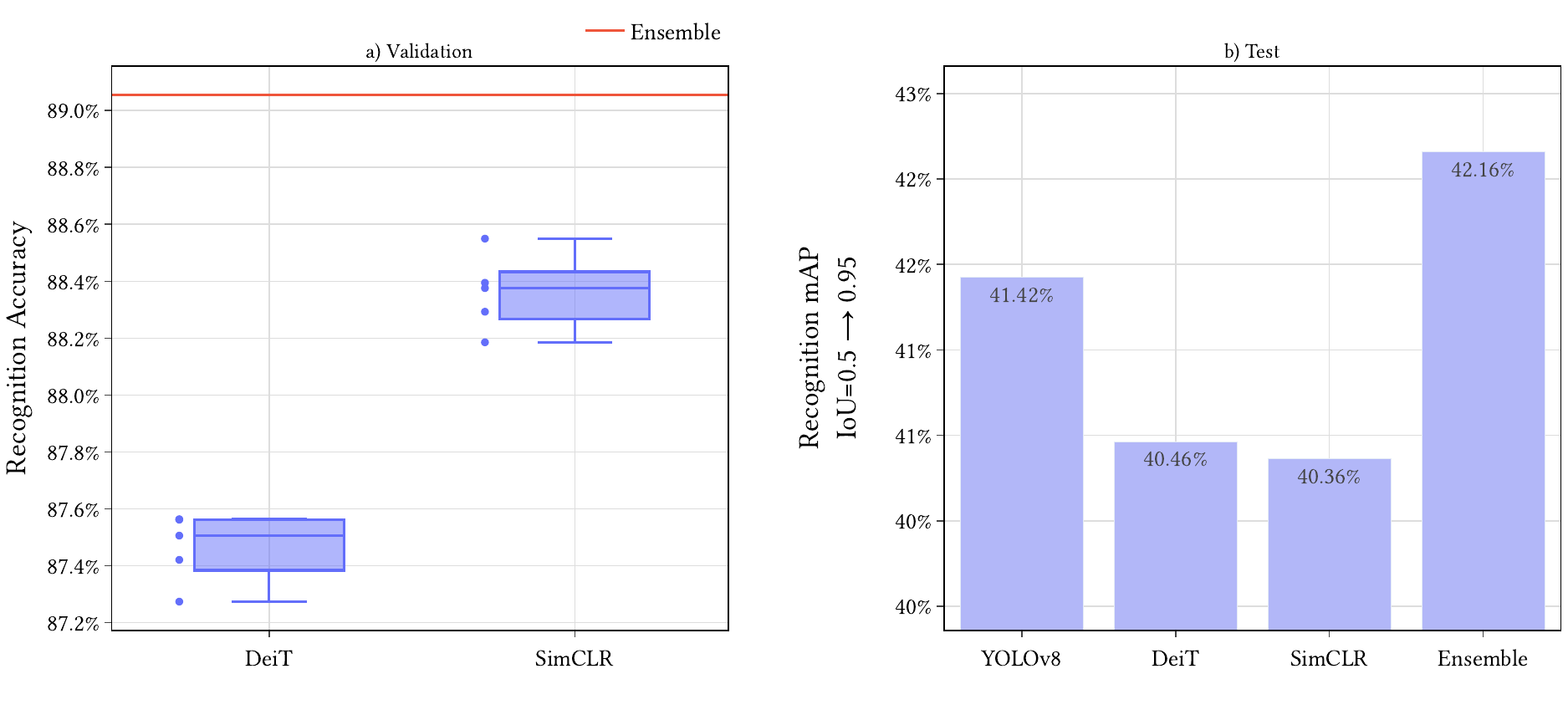}
\caption{Recognition model results. The plot on the left shows accuracy of SimCLR and DeiT models for recognition of cropped characters on the validation partitions. The plot on the right shows the mAP on the test dataset. The YOLOv8 model ensembles of all resolutions from fig. \ref{fig:yolov8-results}c.}
\label{fig:recognition-models}
\end{figure*}

The results of the models trained specifically for recognition are shown in fig. \ref{fig:recognition-models}. The YOLOv8 ensemble outperformed both the DeiT and the SimCLR models. But when they were ensembled, we achieved the highest recognition mAP score of 42.16\%. This was our ultimate submission for the competition. The ensembling process adds new bounding boxes and modifies their confidence scores to account for each of the characters predicted to be present, which results in the slightly lower detection mAP  of 51.42\% for a larger gain in the recognition mAP.

% \begin{figure*}[htbp]
% \includegraphics[width=\textwidth]{figs/confusion-matrix.pdf}
% \caption{Confusion matrix for character predictions on the AL-PUB and competition datasets, ensembling the SimCLR Resnet-50 and DeiT models. Results shown are the predictions obtained using the averaged probability from both methods on validation sets obtained using fivefold cross-validation.}
% \label{fig:confusion_matrix}
% \end{figure*}

% \begin{figure}[htbp!]
% \includegraphics[width=\columnwidth]{figs/RecognitionAccuracy.pdf}
% \caption{Accuracy of SimCLR and DeiT models for recognition of cropped characters.}
% \label{fig:recognition_accuracy}
% \end{figure}

We found that the SimCLR ResNet-50 and DeiT-Small models performed similarly in identifying the Greek character present in an image. Averaging the cross-validation results, we obtained 88.4\% accuracy for the former and 87.4\% accuracy for the latter (fig. \ref{fig:recognition-models}a). The combination of these models achieved 89.1\% accuracy.

The supplementary material contains a confusion matrix for our ensembled predictions, which shows that we obtained reasonable recall for every character, even those poorly represented in the dataset. The most common mistakes are between characters having a similar appearance, such as A and $\Lambda$ or Z and $\Xi$.

\subsection{Competition Results}

\begin{figure*}[ht!]
\includegraphics[width=\textwidth]{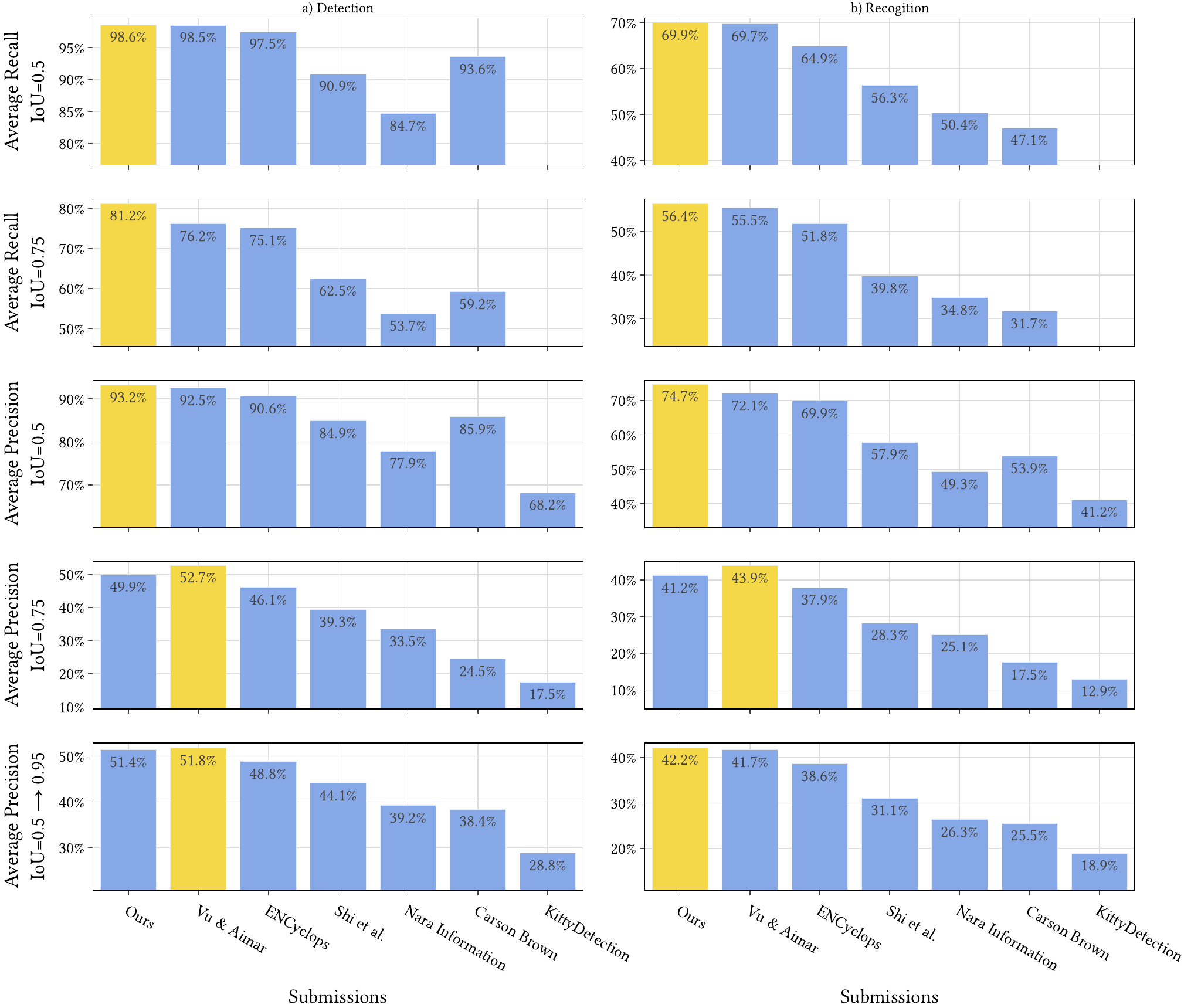}
\caption{Competition results. The best submission for each metric is shown in yellow. The average recall results for KittyDetection were not available.}
\label{fig:competition}
\end{figure*}

The results of our submission in the detection task is shown in fig. \ref{fig:competition}a. In the primary metric (and at an IoU of 0.75) our submission was behind that of Vu \& Aimar, giving us the second place for this task. At the lower IoU value of 0.5, we achieved a much higher mAP at 93.2\%. This was the highest of all submissions. This indicates that while our bounding boxes were not as localized as the winning submission, our method was still able to accurately identify the vast majority of letters. In common with the other submissions, the recall scores were substantially better than the precision scores. We achieved an average recall of 98.6\% at an IoU of 0.5 and 81.2\% at an IoU of 0.75, which were the best of all submissions.

% \begin{figure}[ht!]
% \includegraphics[width=\columnwidth]{figs/Classification-ResultsFull.pdf}
% \caption{Competition recognition results. The best submission for each metric is shown in yellow. The average recall results for KittyDetection were not available.}
% \label{fig:recognition-competition}
% \end{figure}

The results of our submission in the recognition task is shown in fig. \ref{fig:competition}b. In the primary metric we achieved a mAP of 42.2\%, which was the winning submission for this task. As with detection, the result for an IoU of 0.75 was not as good as the submission by Vu \& Aimar but at an IoU of 0.5 our result was 74.7\%, more than 2\% higher than the next closest submission. Our submission also ranked highest on average recall for both an IoU at 0.5 and 0.75.

\section{Oxyrhynchus Papyri Predictions}

\begin{figure*}[ht!]
\includegraphics[width=\textwidth]{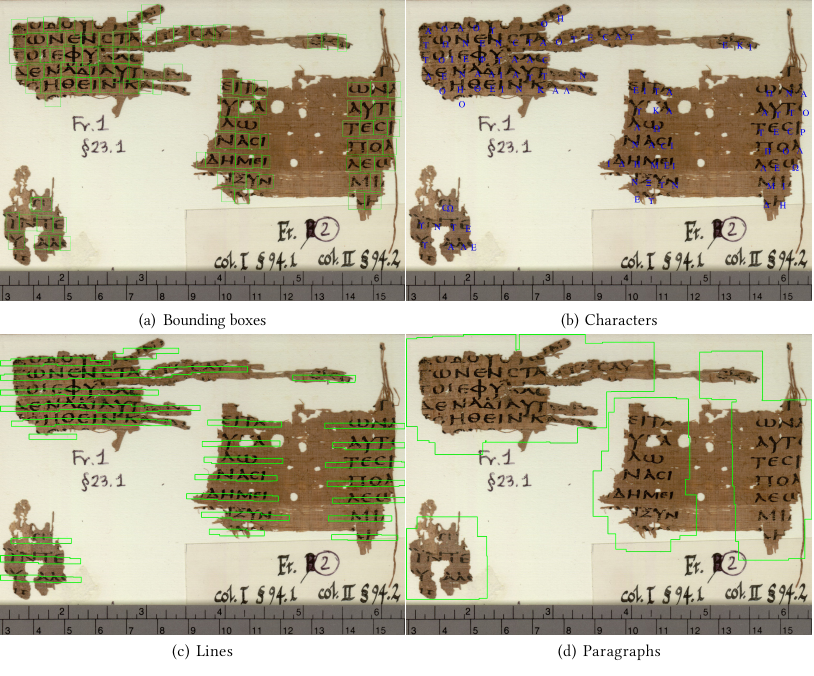}
     \centering
     % \begin{subfigure}[b]{0.5\textwidth}
     %     \centering
     %     \includegraphics[width=\textwidth]{figs/VLVII_n3895/MicrosoftTeams-image.jpeg}
     %     \caption{Bounding boxes}
     %     \label{fig:n3895bounding_boxes}
     % \end{subfigure}%
     % \begin{subfigure}[b]{0.5\textwidth}
     %     \centering
     %     \includegraphics[width=\textwidth]{figs/VLVII_n3895/MicrosoftTeams-image(3).jpeg}
     %     \caption{Characters}
     %     \label{fig:n3895characters}
     % \end{subfigure}
     % \begin{subfigure}[b]{0.5\textwidth}
     %     \centering
     %     \includegraphics[width=\textwidth]{figs/VLVII_n3895/MicrosoftTeams-image(1).jpeg}
     %     \caption{Lines}
     %     \label{fig:n3895lines}
     % \end{subfigure}%
     % \begin{subfigure}[b]{0.5\textwidth}
     %     \centering
     %     \includegraphics[width=\textwidth]{figs/VLVII_n3895/MicrosoftTeams-image(2).jpeg}
     %     \caption{Paragraphs}
     %     \label{n3895paragraphs}
     % \end{subfigure}
        \caption{Model output for papyrus in P. Oxy. LVII.3895 (Thucydides iii 23, 94). Courtesy of The Egypt Exploration Society and the Faculty of Classics, University of Oxford.}
        \label{fig:n3895prediction}
\end{figure*}

To illustrate the potential of our approach to reduce the time required to transcribe manuscripts into a digital format, we applied our model to 4,533 documents from the Oxyrhynchus Papyri collection. Once we obtained the character bounding boxes and predictions for the scanned images of these papyri (as detailed in \S\ref{sec:ensembling}) we removed boxes with low confidence values and, where there was a large degree of overlap between boxes, we retained the box with the highest confidence and discarded the rest. 

We then used feathering of the height and width of the boxes to identify lines of text and paragraphs, allowing us to convert the manuscripts to text. The outputs from this process are shown for a particular manuscript in fig. \ref{fig:n3895prediction}. While more advanced Optical Character Recognition (OCR) approaches might be used for these final steps, this simple approach allows us to conduct keyword searches highlighting the value of these models. For example, a search for the stem of the name Achilles (AXI$\Lambda$*), finds 18 documents, four of which contain the text of the \textit{Iliad} (2748, 3155, 3323, 4817) and in 12 of the remainder (2568, 2672, 2680, 2786, 2960, 2967, 3479, 3774, 4089, 4849, 4991, 5010), the text corresponds to the transcription in the Digital Corpus of Literary Papyri (DCLP) or the Duke Databank of Documentary Papyri (DDbDP) hosted by \url{https://papyri.info}. Only two hits of documentary papyri (2182 and 3250) showed false positives where the DDbDP transcription did not include the query string. Since the original training data came from literary papyri, the results for documentary texts in the Oxyrhynchus Papyri with a cursive script are less reliable than those with a bookhand.

We publicly release the bounding box data using the COCO annotation format, plain text and Text Encoding Initiative (TEI) XML at \url{https://doi.org/10.6084/m9.figshare.25011140.v1}.

\section{Discussion}
\label{sec:discussion}

Since 1898, only a small fraction of the Oxyrhynchus Papyri has been catalogued and transcribed. This work, along with the character recognition models presented by Swindall et al. \cite{swindall2021exploring} and the other submissions to the ICDAR competition \cite{icdar}, demonstrate the potential of new machine learning and AI approaches to make large scale analysis of these documents feasible.

The ability to reliably detect bounding boxes for characters in Greek manuscripts, and to recognize the letters within them, also opens many avenues for quantitative analysis. For example, similarities between the characters from different fragments might be used to discover \textit{disiecta membra} of the same manuscript (such as P.Oxy XVIII.2170 and PSI XI.1218 \citep[54]{poxy18}) or to identify separate manuscripts written by the same scribe (for a list of such instances, see in \cite{johnson2004}, pp. 61--65). 

Similarly, the isolated characters could be used to automatically detect the style of the hand. The isolated characters could be used as input to a machine learning model to predict the approximate date of the manuscript (for a survey of automated techniques for manuscript dating see \cite{omayio2022}). The localization of the characters on the page can automate the process of identifying the widths and heights of the columns, similar to the analysis of \cite{johnson2004}, pp. 162--212. The widths and heights of individual characters could also be used to refine probabilistic models for assessing the content that could have been in manuscript lacunae, such as that of \cite{mccollum2022}.

% Automated Digital Editions

% Attempts have been made to produce large numbers of high quality digital editions such as the work of Institute for Textual Scholarship and Electronic Editing (ITSEE) at the University of Birmingham which have produces hundreds of transcriptions of Greek New Testament manuscripts using the Text Encoding Initiative (TEI) and published these online with a Creative Commons license. This is a massive leap forward. But the task of transcription and quality control are intensely laborious and thus these transcriptions only represent a small proportion of the surviving material in the New Testament corpus. Advances in automated handwritten-

% Cite papyri

% Cite Transkribus\cite{transkribus}

% TEI\cite{ism_tei_1995}\section{Conclusion}
\label{sec:conclusion}

We present the results of our submission to the `ICDAR 2023 Competition on Detection and Recognition of Greek Letters on Papyri'. We used an ensemble of YOLOv8 models to detect and recognize individual characters and refined the recognition predictions by DeiT and SimCLR models. Our predictions were enhanced by supplementary data from the Oxyrhynchus Papyri. Our submission was second in the detection challenge with a mAP for IoU values from 0.5 to 0.95 as 51.42\%, behind the first place winner of 51.83\%. The mAP at an IoU value of 0.5 produced a mAP score of 93.2\% and mAR at 98.6\%, which were the highest of all submissions. For the recognition challenge, our submission produced a mAP of 42.2\% for IoU values from 0.5 to 0.95 and was the winning submission. For the less strict IoU value of 0.5 we achieved the best results with a mAP of 74.7\% and mAR of 69.9\%. We release our predictions on the Oxyrhynchus Papyri in COCO annotation format, plain text, and TEI XML anticipating that others might use these results for further analysis.

\backmatter

\bmhead{Supplementary information}

The confusion matrix from the cross validation results of the recognition specific models discussed in \S\ref{sec:experimental-results} is found in the supplementary material.

\bmhead{Acknowledgements}

We thank David Turnbull, Kamalpreet Singh and Daniel Russo-Batterham for assistance with different aspects of this project. 

\section*{Statements and Declarations}

This research was supported by The University of Melbourne’s Research Computing Services and the Petascale Campus Initiative. This project was undertaken with the assistance of resources and services from the National Computational Infrastructure (NCI), which is supported by the Australian Government. Evelyn Mannix's contribution to this research was supported by an Australian Government Research Training Program (RTP) Scholarship.

\bmhead{Financial interests}
The authors declare they have no financial interests.

\bmhead{Non-financial interests} 
None.

\bmhead{Data availability} 
The training and validation data for the competition is available at \url{https://codalab.lisn.upsaclay.fr/competitions/12419}. Oxyrhynchus Papyri images are available at \url{https://oxyrhynchus.web.ox.ac.uk/images}. The AL-PUB version 2 dataset is available at \url{https://data.cs.mtsu.edu/al-pub/}. Our detections of characters from the Oxyrhynchus Papyri is available at \url{https://doi.org/10.6084/m9.figshare.25011140.v1}.

%%===========================================================================================%%
%% If you are submitting to one of the Nature Portfolio journals, using the eJP submission   %%
%% system, please include the references within the manuscript file itself. You may do this  %%
%% by copying the reference list from your .bbl file, paste it into the main manuscript .tex %%
%% file, and delete the associated \verb+\bibliography+ commands.                            %%
%%===========================================================================================%%

\bibliography{references}% common bib file
%% if required, the content of .bbl file can be included here once bbl is generated
%%\input sn-article.bbl

\end{document}